\documentclass[11pt,a4paper]{article}
\usepackage[hyperref]{acl2017}
\usepackage{times}
\usepackage{latexsym}
\usepackage[utf8]{inputenc} 
\usepackage[T1]{fontenc}    
\usepackage{url}            
\usepackage{booktabs}       
\usepackage{amsfonts}       
\usepackage{nicefrac}       
\usepackage{microtype}      
\usepackage{amsmath}
\usepackage{amsthm}
\usepackage{amssymb}
\usepackage{graphicx}
\usepackage{xspace}
\usepackage{tabularx}
\usepackage{multicol}
\usepackage{multirow}
\usepackage{url}

\newcommand{\bb}{\mbox{\boldmath $b$}}

\newcommand{\bh}{\mbox{\boldmath $h$}}

\newcommand{\bw}{\mbox{\boldmath $w$}}

\newcommand{\by}{\mbox{\boldmath $y$}}
\newcommand{\bz}{\mbox{\boldmath $z$}}

\newcommand{\bW}{\mbox{\boldmath $W$}}

\newcommand{\be}{\begin{eqnarray}}
\newcommand{\ee}{\end{eqnarray}}
\newcommand{\bee}{\begin{eqnarray*}}
\newcommand{\eee}{\end{eqnarray*}}

\newcommand{\matrixb}{\left[ \begin{array}}
\newcommand{\matrixe}{\end{array} \right]}

\usepackage{url}

\aclfinalcopy 

\title{Self-Knowledge Distillation in Natural Language Processing}

\author{Sangchul Hahn \\
  Handong Global University \\ Pohang, South Korea \\
  {\tt schahn21@gmail.com} \\\And
  Heeyoul Choi \\
  Handong Global University \\ Pohang, South Korea \\
  {\tt heeyoul@gmail.com} \\}

\date{}

\begin{document}
\maketitle
\begin{abstract}
Since deep learning became a key player in natural language processing (NLP), many deep learning models have been showing remarkable performances in a variety of NLP tasks, and in some cases, they are even outperforming humans. Such high performance can be explained by efficient knowledge representation of deep learning models. While many methods have been proposed to learn more efficient representation, knowledge distillation from pretrained deep networks suggest that we can use more information from the soft target probability to train other neural networks. In this paper, we propose a new knowledge distillation method \emph{self-knowledge distillation}, based on the soft target probabilities of the training model itself, where multimode information is distilled from the word embedding space right below the softmax layer. Due to the time complexity, our method approximates the soft target probabilities. In experiments, we applied the proposed method to two different and fundamental NLP tasks: language model and neural machine translation. The experiment results show that our proposed method improves performance on the tasks.
\end{abstract}

\section{Introduction}

Deep learning has achieved the state-of-the-art performance on many machine learning tasks, such as image classification, object recognition, and neural machine translation \citep{He2015resnet, Redmon2016yolo9000, Vaswani2017} and outperformed humans on some tasks. In deep learning, one of the critical points for success is to learn better representation of data with many layers \citep{Bengio2014} than other machine learning algorithms. In other words, if we make a model to learn better representation of data, the model can show better performance. 

In natural language processing (NLP) tasks like language modeling (LM) \citep{Bengio2003nnlm,Mikolov2013} and neural machine translation (NMT) \citep{Sutskever2014,Bahdanau2015}, when the models are trained, they are to generate many words in sentence, which is a sequence of classification steps, for each of which they choose a target word among the whole words in the dictionary. That is why LM and NMT are usually trained with the sum of cross-entropies over the target sentence. Thus, although language related tasks are more of generation rather than classification, the models estimate target probabilities with the softmax operation on the previous neural network layers and the target distributions are provided as one-hot representations. As data representation in NLP models, word symbols should also be represented as vectors. 

In this paper, we focus on the word embedding and the estimation of the target distribution. In NLP, word embedding is a step to translate word symbols (indices in the vocabulary) to vectors in a continuous vector space and is considered as a standard approach to handle symbols in neural networks. When two words have semantically or syntactically similar meanings, the words are represented closely to each other in a word embedding space. Thus, even when the prediction is not exactly correct, the predicted word might not be so bad, if the estimated word is very close to the target word in the embedding space like `programming' and `coding'. That is, to check how wrong the prediction is, the word embedding can be used. There are several methods to obtain word embedding matrices \citep{Mikolov2013, Pennington2014glove}, in addition to neural language models \citep{Bengio2003nnlm, Mikolov2010}. Recently, several approaches have been proposed to make more efficient word embedding matrices, usually based on contextual information \citep{Levy2017, hchoi2017csl}.  

On the other hand, knowledge distillation was proposed by \cite{Hinton2015} to train new and usually shallow networks using hidden knowledge in the probabilities produced by the pretrained networks. It shows that there is knowledge not only in the target probability corresponding to the target class but also in the other class probabilities in the estimation of the trained model. In other words, the other class probabilities can contain additional information describing the input data samples differently even when the samples are in the same class. Also, samples from different classes could produce similar distributions to each other. 


In this paper, we propose a new knowledge distillation method, {\em self-knowledge distillation} (SKD) based on the word embedding of the training model itself. That is, self-knowledge is distilled from the predicted probabilities produced by the training model, expecting the model has more information as it is more trained. In the conventional knowledge distillation, the knowledge is distilled from the estimated probabilities of pretrained (or teacher) models. Contrary, in the proposed SKD, knowledge is distilled from the current model in the training process, and the knowledge is hidden in the word embedding. During the training process, the word embedding reflects the relationship between words in the vector space. A word close to the target word in the vector space is expected to have similar distribution after softmax, and such information can be used to approximate the soft target probability as in knowledge distillation. 
We apply our proposed method to two popular NLP tasks: LM and NMT. The experiment results show that our proposed method improves the performance of the tasks. Moreover, SKD reduces overfitting problems which we believe is because SKD uses more information. 


The paper is organized as follows. Background is reviewed in Section 2. In Section 3, we describe our proposed method, SKD. Experiment results are presented and analyzed in Section 4, followed by Section 5 with conclusion.

\section{Background}
In this section, we briefly review the cross-entropy and knowledge distillation. Also, since our proposed method is based on word embedding, the layer right before the softmax operation, word embedding process is summarized. 

\subsection{Cross Entropy}

For classification with $C$ classes, neural networks produce class probabilities $p_i$, $i \in \{0, 1, ... C\}$ by using a softmax output layer which calculates class probabilities from the logit, $z_i$ considering the other logits as follows.
\begin{equation}
p_i = \frac{\exp{(z_i)}}{\sum_k \exp{(z_k)}}. 
\label{eq:softmax}
\end{equation}

In most classification problems, the objective function for a single sample is defined by the cross-entropy as follows. 
\be
J(\theta) &=& -\sum_k y_k \log p_k,
\label{eq:ce_all}
\ee
where $y_k$ and $p_k$ are the target and predicted probabilities. The cross-entropy can be simply calculated by 
\be
J(\theta) &=& - \log p_t,
\label{eq:ce}
\ee
when the target probability $\by$ is a one-hot vector defined as
\begin{align}
    \label{eq:one_hot}
    y_k = \left\{ 
        \begin{array}{l l}
            1,&\text{ if }k = t (\mbox{target class}) \\
            0,&\text{ otherwise}
        \end{array}
        \right..
\end{align} 

Note that the cross-entropy objective function says only how likely input samples belong to the corresponding target class, and it does not provide any other information about the input samples.

\subsection{Knowledge Distillation}

A well trained deep network model contains meaningful information (or knowledge) extracted from training datasets for a specific task. Once a deep model is trained for a task, the trained model can be used to train new smaller (shallower or thinner) networks as shown in \cite{Hinton2015,Romero2014fitnet}. This approach is referred to as {\em knowledge distillation}. 

Basically, knowledge distillation provides more information to new models for training and improves the new model's performance. Thus, when a new model which is usually smaller is trained with the distilled knowledge from the trained deep model, it can achieve a similar (or sometimes even better) performance compared to the pretrained deep model. 

In the pretrained model, knowledge lies in the class probabilities produced by softmax of the model as in Eq. (\ref{eq:softmax}). All probability values including the target class probability describe relevant information about the input data. Thus, instead of one-hot representation of the target label where only the target class is considered in cross-entropy, all probabilities over the whole classes from the pretrained model can provide more information about the input data in cross-entropy, and can teach new models more efficiently. All probabilities from the pretrained model are considered as {\em soft target probabilities}. 

In a photo tagging task, depending on the other class probabilities, we understand the input image better than just target class. When a class `mouse' has the highest probability, if `mascot' has a relatively high probability, then the image would be probably `mickey mouse'. If `button' or `pad' has a high probability, the image would be a mouse as a computer device. The other class probabilities have some extra information and such knowledge in the pretrained model can be transferred to a new model by using a soft target distribution of the training set. 

When the target labels are available, the objective function is a weighted sum of the conventional cross-entropy with the correct labels and the cross-entropy with the soft target distribution, given by 
\begin{equation}
J(\theta) = -(1-\lambda) \log p_t  - \lambda \sum_k q_k \log p_k, 
\label{eq:obj}
\end{equation}
where $p_k$ is probability for class $k$ produced by current model with parameter $\theta$, and $q_k$ is the soft target probability from the pretrained model. $\lambda$ controls the amount of knowledge from the trained model. Note that the conventional knowledge distillation extracts knowledge from a pretrained model, and in this paper, we propose to extract knowledge from the current model itself without any pretrained model.

Furthermore, in a recently proposed paper by \citep{born_again}, they proved that knowledge distillation can be useful to train a new model which has the same size and the same architecture as the pretrained model. They trained a teacher model first, then they trained a student model with distilled knowledge from the teacher model. Their experiment results show that the student models outperform the teacher model. Also, even though when the teacher model has a less powerful architecture, the knowledge from the trained teacher model can boost student models which have more powerful (or bigger) architectures. It means that even the knowledge is distilled from a relatively weak model, it can be useful to train a bigger model.

\subsection{Word Embedding}
Word embedding is to convert symbolic representation of words to vector representation with semantic and syntactic meanings, which reflects the relations between words. Including CBOW, Skip-gram \citep{Mikolov2013}, and GloVe \citep{Pennington2014glove}, various word embedding methods have been proposed to learn a word embedding matrix. The trained embedding matrix can be transferred to other models like LM or NMT \cite{Ahn2016}.

CBOW predicts a word given its neighbor words, and Skip-gram predicts the neighbor words given a word. They use feedforward layers, and the last layer of CBOW includes the word embedding matrix, $\bW$, as follows. 
\begin{equation}
\bz = \bW \bh + \bb, 
\label{eq:ff}
\end{equation}
where $\bb$ is a bias, $\bh$ is hidden layer, and $\bz$ is logits for the softmax operation. 

\begin{figure}[t]
\centerline{\hbox{
\includegraphics[height=1.5in]{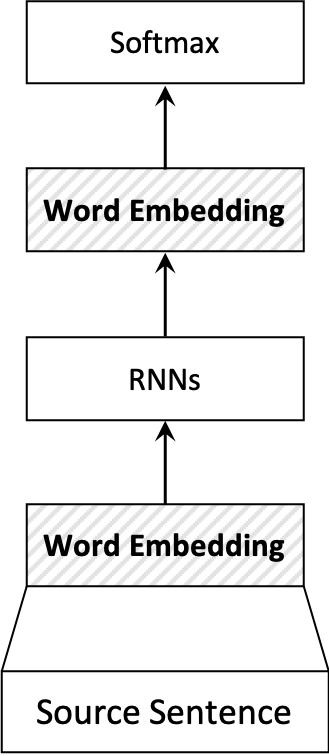}
\hspace{0.5in}
\includegraphics[height=2.4in]{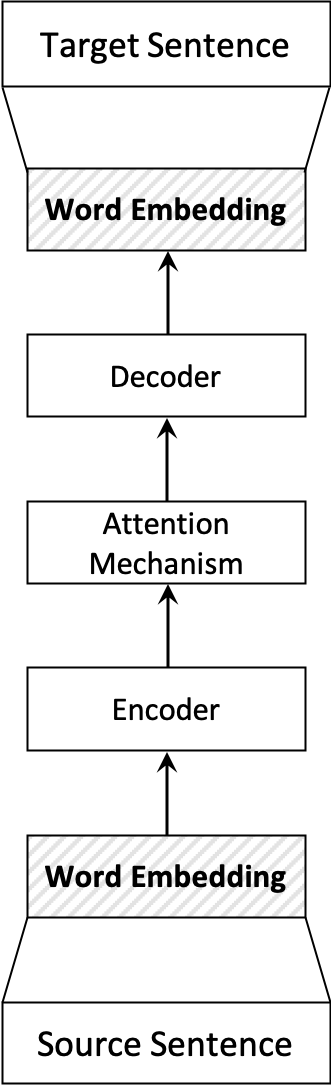}
}}
\centerline{\hbox{ 
(a) Language Model  \hspace{0.37in} (b) NMT Model \hspace{0.15in}  }}
\caption{Network architectures of LM and NMT. Word embedding is presented as gray boxes in the models.}
\label{fig:skd_embedding}
\end{figure}

Words in the embedding space have semantic and syntactic similarities, such that two similar words are close in the space. Thus, when the classification is not correct, the error can be interpreted differently depending on the similarity between the predicted word and the target word.  For example, when the target word is `learning', if the predicted word is `training', then it is less wrong than other words like `flower' or `internet'. In this paper, we utilize such hidden information (or knowledge) in the word embedding space, while training. 
Fig. \ref{fig:skd_embedding} shows where the word embedding is located in LM and NMT, respectively.

\section{Self-Knowledge Distillation}

We propose a new learning method {\em self-knowledge distillation} (SKD) which distills knowledge from a currently training model, following the conventional knowledge distillation. In this section, we describe an algorithm for SKD and its application to language model and neural machine translation.

\subsection{SKD Equations}

In order to apply knowledge distillation on a current training model, we need to obtain soft target probabilities as $q_k$ in Eq. (\ref{eq:obj}) for all classes, but they are not available explicitly. However, when the model is trained enough, then the word embedding has such information implicitly. If a word $\bw_i$ is close to $\bw_j$ in the embedding space, the probability $p_i$ would be close to $p_j$ for a given input sample.

When $t$ is the target class, we calculate the soft target probabilities $q_k$ based on the word embedding. First, we assume that $q_t$ should be high, and if $\bw_k$ is close to $\bw_t$ in the embedding space, $q_k$ should be also high. That is, the Euclidean distance between words is used to estimate the soft target probability. The other class probabilities (or soft target probabilities) $q_k$ can be obtained by 
\begin{equation}
q_k = \frac{1}{Z} \exp\{-\sigma \| \bw_t - \bw_k \|_{2}\}, 
\label{eq:qk1}
\end{equation}
where $\| \cdot \|_{2}$ is $l2$-norm, and $Z$ is a normalization term. $\sigma$ is a scale parameter and its value depends on the average distance to the corresponding nearest neighbors in the word embedding space. 
However, due to the expensive computational cost, we do not calculate $q_k$ for all classes, and we choose just one of the other classes, which is the predicted class of the current model. 

\begin{figure}[ht]
\centerline{\hbox{ \includegraphics[width=2.5in]{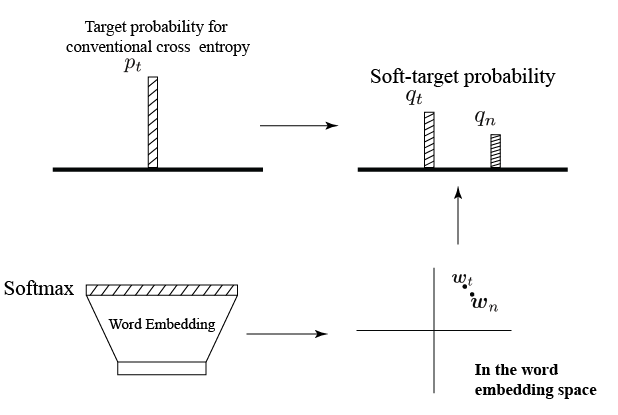}}}
\caption{Given a target class $t$, a soft target probabilities are obtained based on the distance in the word embedding space. However, only the target class and the predicted class have soft target probabilities in SKD.}
\label{fig:skd_overview}
\end{figure}

Assuming that the model predicts a class $n$ for a given input sample, only $q_t$ and $q_n$ are used as distilled knowledge. We clip the $q_n$ value with 0.5, meaning that the class $n$ cannot be more correct than the real target $t$, so Eq. (\ref{eq:qk1}) becomes 
\be
q_n &=& \min\{\exp\{-\sigma \| \bw_t - \bw_n \|_{2}\}, 0.5\},\nonumber \\ 
q_t &=& 1-q_n,
\label{eq:qk}
\ee
where $q_n + q_t = 1$. That is, we consider only two soft target probabilities as shown in Fig. \ref{fig:skd_overview}. Note that we use Euclidean distance between $\bw_t$ and $\bw_n$ to calculate $q_n$, but other approaches like inner product would be possible. 

Now, the objective function of SKD becomes similar to Eq. (\ref{eq:obj}), and is defined by 
\be
J(\theta) &=& -(1-\lambda) \log p_t \nonumber \\
&& - \lambda (q_t \log p_t  + q_n \log p_n),
\label{eq:obj_skd}
\ee
where the second term of Eq. (\ref{eq:obj}) is approximated by $\lambda (q_t \log p_t  + q_n \log p_n)$, ignoring the other class probabilities. Eq. (\ref{eq:obj_skd}) can be rewritten simply as follows.
\be
J(\theta) = -(1-\lambda q_n) \log p_t  - \lambda q_n \log p_n. 
\label{eq:obj_skd_simple}
\ee

Eqs. (\ref{eq:obj_skd}) or (\ref{eq:obj_skd_simple}) can be understood in three cases. First, if the prediction is correct ($n = t$), then Eq. (\ref{eq:obj_skd}) is the same as the conventional cross-entropy objective.  Second, if $\bw_n$ is far from $\bw_t$ in the word embedding space, then $q_n$ is close to zero and Eq. (\ref{eq:obj_skd}) becomes close to the conventional cross-entropy objective. Finally, if $\bw_n$ is close to $\bw_t$ (e.g. $q_n$ = 0.4), it approximates the soft target probability with only two classes $t$ and $n$, and the model is trained to produce probabilities for class $t$ and $n$ as close as $q_t$ and $q_n$. This approach trains the model with different targets for different input samples. 

Fig. \ref{fig:skd_overview} presents how SKD obtains simplified soft target distribution based on the distance of target and estimated vectors in the word embedding space. 

\subsection{SKD Algorithm}

Since SKD distills knowledge from the current training model, at the beginning of the training process, the model does not contain relevant information. That is, we cannot extract any knowledge from the training model at the beginning. Thus, we start training process without knowledge distillation at first and gradually increase the amount of knowledge distillation as the training iteration goes. So, our algorithm starts with the conventional cross-entropy objective function in Eq. (\ref{eq:ce}), and after training the model for a while, it gradually transits to Eq. (\ref{eq:obj_skd_simple}). To implement the transition, another parameter $\alpha$ is introduced to Eq. (\ref{eq:obj_skd_simple}), leading to the final objective function as follows.
\be
J(\theta) = -(1-\alpha \lambda q_n ) \log p_t  - \alpha \lambda q_n \log p_n, 
\label{eq:obj_skd2}
\ee
$\alpha$ starts from 0 with which Eq. (\ref{eq:obj_skd2}) becomes the conventional cross-entropy. After $K$ iterations, $\alpha$ increases by $\eta$ per iteration and eventually goes up to 1 with which Eq. (\ref{eq:obj_skd2}) becomes the same as Eq. (\ref{eq:obj_skd}). In our experiments, we used a simplified equation as in Eq. (\ref{eq:obj_skd3}) without $\lambda$ so that the objective function relies gradually more on the soft target probabilities as training goes. 
\be
J(\theta) = -(1-\alpha q_n ) \log p_t  - \alpha q_n \log p_n. 
\label{eq:obj_skd3}
\ee
Table \ref{table:skd_algo} summarizes the proposed SKD algorithm. 

\begin{table}[ht]
\caption{Self-Knowledge Distillation Algorithm}
\label{table:skd_algo}
  \centering
  \begin{tabular}{lll}
    \toprule
    Algorithm 1: SKD Algorithm \\
    \midrule
    Initialize the model parameters $\theta$ \\
    Initialize $\alpha = 0$ and $\sigma$ \\
    (See the experiments for $\sigma$ values.) \\
    Repeat $K$ times: \\
    ~~~~~ Train the network based on the\\
    ~~~~~ cross-entropy in Eq. (\ref{eq:ce}) \\
    Repeat until convergence: \\
    ~~~~~ Train the network based on \\
    ~~~~~ the SKD objective function in Eq. (\ref{eq:obj_skd3}) \\
    ~~~~~ Update $\alpha$ with $\alpha + \eta$  \\
    ~~~~~ (See the experiments for $\eta$ values.) \\
    \bottomrule
\end{tabular}
\end{table}

\subsection{NLP Tasks}

SKD is applied to two different NLP tasks: language modeling (LM), and neural machine translation (NMT). Although LM and NMT are actually sentence generation rather than classification, they have classification steps to generate words for the target sentence. Also, the sum of cross-entropies over the words in the sentence is adapted as an objective function for them. 

In addition, to check if SKD is robust against errors in the word embedding space, we also evaluate SKD when we add Gaussian noise in the word embedding space for target words in the decoder.

\section{Experiments}
To evaluate self-knowledge distillation, we compare it to the baseline models for language modeling and neural machine translation.

\subsection{Dataset}
For language modeling, we use two different datasets: Penn TreeBank (PTB) and WiKi-2. PTB was made by \citep{Marcus1993penntree}, and we use the pre-processed version by \citep{Mikolov2010}. In the PTB dataset, the train, valid and test sets have about 887K, 70K, and 78K tokens, respectively, where the vocabulary size is 10K. The WiKi-2 dataset introduced by \citep{merity2016} consists of sentences that are extracted from Wikipedia. It has about 2M, 217K, and 245K tokens for train, valid, and test sets. Its vocabulary size is about 33K.
We did not apply additional pre-processing for the PTB dataset. The WiKi-2 dataset is pre-tokenized data, therefore we only added an end-of-sentence token (<EOS>) to every sentence. 

For machine translation, we evaluated models on three different translation tasks (En-Fi, Fi-En, and En-De) with the available corpora from WMT'15 \footnote{http://www.statmt.org/wmt15/translation-task.html}. The dictionary size is 10K for En-fi and Fi-En translation task, and 30K for the En-De translation task.

\subsection{Language Modeling}
Language modeling (LM) has been used in many different NLP tasks like automatic speech recognition (ASR), and machine translation (MT) to capture syntactic and semantic structure of a natural language. The neural network-based language models (NNLM) and recurrent neural network language model (RNNLM) catch the syntactic and semantic regularities of an input language \citep{Bengio2003nnlm,Mikolov2013}. RNNLM is our baseline, which consists of a single LSTM layer and single feed forward layer with ReLU \citep{Le2015}.

We evaluate four models: Baseline, Noise (with Gaussian noise on the word embedding), SKD, and Noise+SKD. To show that the information by SKD is more knowledgeable than random noise, we tested a noise injected model which injects only Gaussian noise to the word embedding space. The word dimension is set to 500 and the number of hidden nodes is 400 for all models. We set the $\sigma$ and $\eta$ in the SKD algorithm in Table \ref{table:skd_algo} 0.1 (both PTB and WiKi-2 dataset) and 0.0002 (PTB), 0.00011 (WiKi-2), respectively. We applied the SKD object function after 500 batches for PTB and 900 batches for WiKi-2. Note that Wiki-2 data is larger than PTB. 

The evaluation metric is the negative log-likelihood (NLL) for each sentence (the lower is the better). Table \ref{table:lm_perp} presents NLLs for the test data of two datasets with different models. It shows that our proposed methods (both noise injection and self-distillation knowledge) improve the results in the LM task. Note that SKD provides more knowledgeable information than Gaussian noise.

\begin{table}[ht]
\caption{NLLs for LM with different models on PTB and Wiki-2.}
\label{table:lm_perp}
  \centering
  \begin{tabular}{|l|r|r|}
    \hline
    Model & PTB & Wiki-2 \\
    \hline
    Baseline & 101.40 & 119.49 \\
    +Noise & 101.28 & 118.70 \\
    +SKD & 99.38 & 116.85 \\
    +Noise+SKD & \textbf{97.41} & \textbf{116.60} \\
\hline
\end{tabular}
\end{table}

\subsection{Neural Machine Translation}
NMT has been widely used in machine translation research, because of its powerful performance and end-to-end training \citep{Sutskever2014,Bahdanau2015,Wu2016}. Attention-based NMT models consist of an encoder, a decoder, and the attention mechanism \citep{Bahdanau2015}, which is our baseline in this paper except for replacing GRU with LSTM and using BPE \cite{Sennrich2016bpe}. The encoder takes the sequence of source words in the word embedding form. The decoder works in a similar way to LM, except the attention mechanism. See \cite{Bahdanau2015} for NMT and the attention mechanism in detail. 

In the experiments, we check how much SKD can improve model's performance using the simple baseline architecture. Since SKD modifies only the objective function, we believe that the improvement by SKD is regardless of model architectures. 

Table \ref{table:nmt_bleu} shows that our proposed method improves NMT performance by around 1 BLEU score. For qualitative comparison, some translation results are presented below. The overall quality of translation of the SKD model looks better than baseline model's. In other words, when the BLEU scores are similar, the sentences translated by the SKD model look better. 

\begin{itemize}
	\item \small(src) Hallituslähteen mukaan tämä on yksi monista ehdotuksista, joita harkitaan.\\
        \small(trg) A governmental source says this is one of the many proposals to be considered.\\
        (baseline)\emph{  According to government revenue, this is one of the many proposals that are being considered to be considered.}\\
        (SKD)\emph{ According to the government, this is one of the many proposals that are being considered.}\\
    \item \small(src) Meillä on hyvä tunne tuotantoketjunvahvuudesta. \\
        \small(trg) We feel very good about supply chain capability. \\
        (baseline)\emph{ We have good knowledge of the strength of the production chain.} \\
        (SKD)\emph{ We have a good sense of the strength of the production chain.} \\
    \item \small(src) En ole oikein tajunnut, että olen näin vanha. \\
        \small(trg) I haven’t really realized that I’m this old. \\
        (baseline)\emph{  I have not been right to realise that I am so old.} \\
        (SKD)\emph{ I am not quite aware that I am so old.} \\
    \item \small(src) Ne vaikuttavat vasta tulevaisuudessa. \\
        \small(trg) They’ll have an impact in the future only. \\
        (baseline)\emph{ They will only be affected in the future.} \\
        (SKD)\emph{ They will only affect in the future.} \\
    
\end{itemize}

\begin{figure}[h!]
\centerline{\hbox{ 
\includegraphics[height=1.2in]{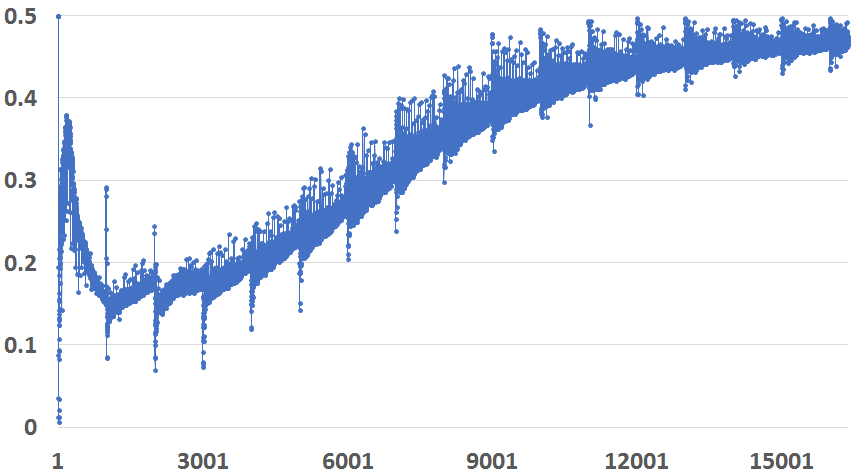}
}}
\centerline{\hbox{ (a) $q_n$ value during NMT model training}}
\vspace{0.1in}
\centerline{\hbox{ 
\includegraphics[height=1.2in]{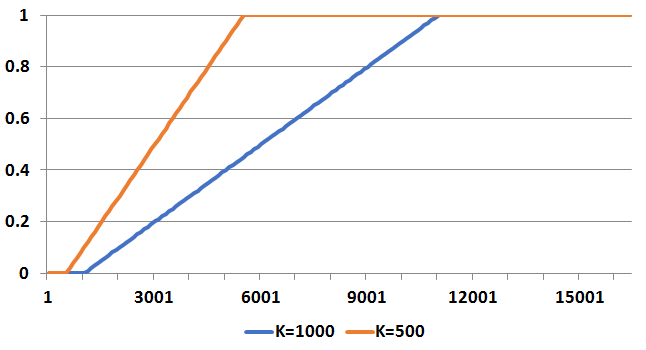}
}}
\centerline{\hbox{ 
(b) Scheduling of $\alpha$ value of NMT training}}
\caption{(a) Change of $q_n$ value during NMT model training for En-Fi translation task, and (b) scheduling of $\alpha$ value in Eq. (\ref{eq:obj_skd3}) of NMT training for En-Fi translation task. (a) shows that when the model is trained more, the $q_n$ value become more close to the target.}
\label{fig:qt_alpha_curve}
\end{figure}

Fig. \ref{fig:qt_alpha_curve} shows a trajectory of the $q_n$ values and scheduling of the $\alpha$ value during training the En-Fi NMT model described in Eq. (\ref{eq:obj_skd3}), respectively. As expected, the $q_n$ value becomes larger than $0.5$ which means that $w_n$ (the predicted word vector) is close enough to the $w_t$ (the target word vector). Fig. \ref{fig:qt_alpha_curve}(b) shows the scheduled value of $\alpha$ in Eq. (\ref{eq:obj_skd3}). The $\alpha$ value starts from $0$ and increases up to $1$ while training. The model is trained with only the cross-entropy for $K$ iterations, and then when the model captures enough knowledge to be distilled, $\alpha$ increases to utilize knowledge from the model.  

\begin{figure}[ht!]
\centerline{\hbox{ \includegraphics[height=1.5in]{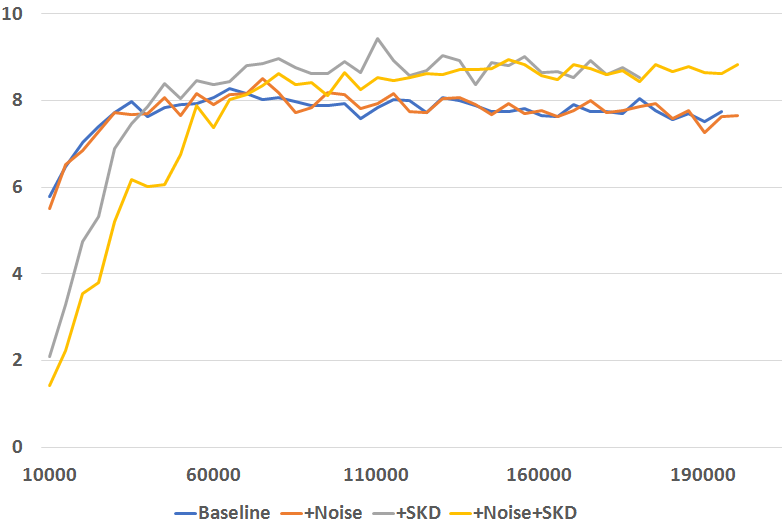}}}
\caption{BLEU scores of validation data while training on En-Fi corpus with four different models: Baseline, +Noise, +SKD, and +Noise+SKD. The vertical axis indicates BLEU score and the horizontal axis the number of training iteration.}
\label{fig:skd_training_curve}
\end{figure}

Also, as in Fig. \ref{fig:skd_training_curve}, the SKD models are not (or more slowly) overfitted to the training data. We believe that SKD provides more information distilled by the training model itself to prevent overfitting. Note that there is no significant difference in the improvements by SKD and Noise, but Noise+SKD improves further. It implies that SKD provides different kinds of information from noise, while the synergy effect between SKD and noise needs more research. 

\begin{table}[!ht]
\caption{BLEU scores on the test sets for En-Fi, Fi-En and En-De with two different beam widths. The scores on the development sets are in the parentheses.}
\label{table:nmt_bleu}
  \centering
  \begin{tabular}{|l|p{2.0 cm}|p{2.0 cm}|}
  \hline
\multicolumn{1}{|c|}{\multirow{2}{*}{\small Model}} & \multicolumn{2}{c|}{\small Beam width} \\ 
\cline{2-3} \multicolumn{1}{|c|}{} & ~~~~~~~~ \small 1 & ~~~~~~~ \small 12\\ \hline \hline
\multicolumn{3}{|c|}{\small En-Fi}      \\ \hline 
\small Baseline     & ~~\small 7.29(8.28)     & ~~\small 9.01(9.85)     \\ \hline
\small +Noise       & ~~\small 7.68(8.50)     & ~~\small 9.35(9.53)     \\ \hline
\small +SKD         & ~~\small 8.36\textbf{(9.43)}     & ~\small 9.87(10.30)    \\ \hline
\small +Noise+SKD   & ~~\small \textbf{8.81}(8.95) & \small \textbf{10.13(10.47)} \\ \hline \hline
\multicolumn{3}{|c|}{\small Fi-En} \\\hline
\small Baseline     & \small 10.42(11.39)   & \small 11.89(12.78)   \\ \hline
\small +Noise       & \small 10.74(11.80)   & \small 12.39(13.35)   \\ \hline
\small +SKD         & \small 10.70(12.52)   & \small 12.43(13.82)   \\ \hline
\small +Noise+SKD   & \small \textbf{11.87(12.92)} & \small \textbf{13.16(14.13)} \\ \hline \hline
\multicolumn{3}{|c|}{\small En-De}      \\ \hline
\small Baseline     & \small 19.72(19.28)   & \small 22.25(20.91)   \\ \hline
\small +Noise       & \small 20.69(19.68)   & \small 22.40(20.92)   \\ \hline
\small +SKD         & \small 20.29(20.41)   & \small 22.59(21.75)   \\ \hline
\small +Noise+SKD   & \small \textbf{21.16(20.34)} & \small \textbf{23.07(21.64)} \\ \hline
\end{tabular}
\end{table}

\section{Conclusion}
We proposed a new knowledge distillation method, self-knowledge distillation, from the probabilities of the currently training model itself. The method uses only two soft target probabilities that are obtained based on the word embedding space. The experiment results with language modeling and neural machine translation show that our method improves the performance. This method can be straightforwardly applied to other tasks where the cross-entropy is used.

As future works, we want to apply SKD to other applications with different model architectures, to show that SKD does not depend on tasks nor the model architectures. For image classification tasks, if we abuse the term `word embedding' to refer to the layer right before the softmax operation, it may be possible to apply SKD in a similar way, although it is not guaranteed that comparable image classes are closely located in the word embedding space for image related tasks. Also, we can develop an automatic way for the parameters like $\alpha$ in Eq. (\ref{eq:obj_skd3}), and generalize the equation for $q_n$ in Eq. (\ref{eq:qk}).

\section*{Acknowledgement}
This research was supported by Basic Science Research Program through the National Research Foundation of Korea(NRF) funded by the Ministry of Education (2017R1D1A1B03033341), and by Institute for Information \& communications Technology Promotion(IITP) grant funded by the Korea government(MSIT) (No. 2018-0-00749, Development of virtual network management technology based on artificial intelligence).





\bibliography{hchoi2018}
\bibliographystyle{acl_natbib}

\end{document}